%% file: 00_root.tex
\documentclass[letterpaper, 10 pt, conference]{ieeeconf}  

\IEEEoverridecommandlockouts                              

\usepackage{graphics} 
\usepackage{epsfig}
\usepackage{amsmath} 
\usepackage{amssymb}  
\usepackage{cite}
\usepackage{balance}
\usepackage{color}

\newcommand{\lf}[1]{{}^{\scriptscriptstyle {#1}}\hspace{-0.1mm}}
\newcommand{\lfR}[1]{{}^{\scriptscriptstyle {#1}}\hspace{-0.8mm}R}

\newcommand{\rf}[1]{_{\scriptscriptstyle {#1}}\hspace{-0.0mm}}
\newcommand{\lfdotR}[1]{{}^{\scriptscriptstyle {#1}}\hspace{-0.8mm}\dot{R}}

\newcommand{\vect}[1]{\boldsymbol{\mathbf{#1}}} 
\newcommand\numberthis{\addtocounter{equation}{1}\tag{\theequation}} 

\DeclareMathOperator*{\Exp}{\mathrm{Exp}}
\DeclareMathOperator*{\Ln}{\mathrm{Ln}}

\title{\LARGE \bf
Best Axes Composition: Multiple Gyroscopes IMU Sensor Fusion to Reduce Systematic Error
}

\author{Marsel Faizullin \and Gonzalo Ferrer%
\thanks{The authors are with Skolkovo Institute of Science and Technology.
 {\tt\small \{marsel.faizullin, g.ferrer\}@skoltech.ru}
\newline 978-1-6654-1213-1/21/\$31.00 \textcopyright 2021 IEEE
}
}

\begin{document}

\maketitle
\thispagestyle{empty}
\pagestyle{empty}

\begin{abstract}
In this paper, we propose an algorithm to combine multiple cheap Inertial Measurement Unit (IMU) sensors to calculate 3D-orientations accurately.
Our approach takes into account the inherent and non-negligible systematic error in the gyroscope model and provides a solution based on the error observed during previous instants of time.
Our algorithm, the {\em Best Axes Composition} (BAC), chooses dynamically the most fitted axes among IMUs to improve the estimation performance.
We compare our approach with a probabilistic Multiple IMU (MIMU) approach, and we validate our algorithm in our collected dataset.
As a result, it only takes as few as 2 IMUs to significantly improve accuracy, while other MIMU approaches need a higher number of sensors to achieve the same results.
\end{abstract}

\input{01_introduction}
\input{02_background}
\input{03_method}
\input{04_experiments}
\input{05_conclusion}


\bibliographystyle{bib/IEEEtran}
\balance
\bibliography{bib/IEEEabrv,bib/imu}

\end{document}

%% file: 01_introduction.tex
\section{Introduction}\label{sec_intro}

Inertial Measurement Unit (IMU) sensors provide information regarding the angular velocity and acceleration of moving objects and are widely used in many applications, 
such as pose estimation, aviation, camera stabilization, virtual reality, visual-inertial odometry, etc. 
One of the reasons why they are so popular is because IMU complements other sensing modalities, such as cameras or Lidars which provide a slow but global observation, by providing a high-rate stream of data.

Open-loop conditions are met when IMU state estimation is not supported by any complementary sensor.
The aim of the current work is to provide the most accurate orientation estimation during this lapse of time. Many applications, such as
high dynamic motion by aerial robots~\cite{scaramuzza2019visual} or augmented reality equipment~\cite{calloway2020three} require virtually no delay between real motions and their estimations. Visual odometry algorithms add a delay of several tens of milliseconds. In contrast, IMU sensors provide high data rate of measurements (up to 1 kHz and higher) at almost negligible processing time.



Microelectromechanical systems (MEMS) IMU are affordable sensors of limited quality. Thereby, multiple of these sensors should improve their accuracy. However, the problem of fusing multiple IMU data is usually tackled as a probabilistic approach given its random nature.
MEMS IMUs produce additional sources of error that are not explained by common models, such as non-linearity~\cite{wang2019optimized}, sensor finite dimensions~\cite{rehder2016extending}, discrete integration, non-Gaussian distributions, non-stationary parameters, temperature dependence, power supply conditions, etc.
These factors or errors are simply denoted as {\em systematic} errors or {\em epistemic} errors.


\begin{figure}[t]
\frame{\includegraphics[width=\columnwidth]{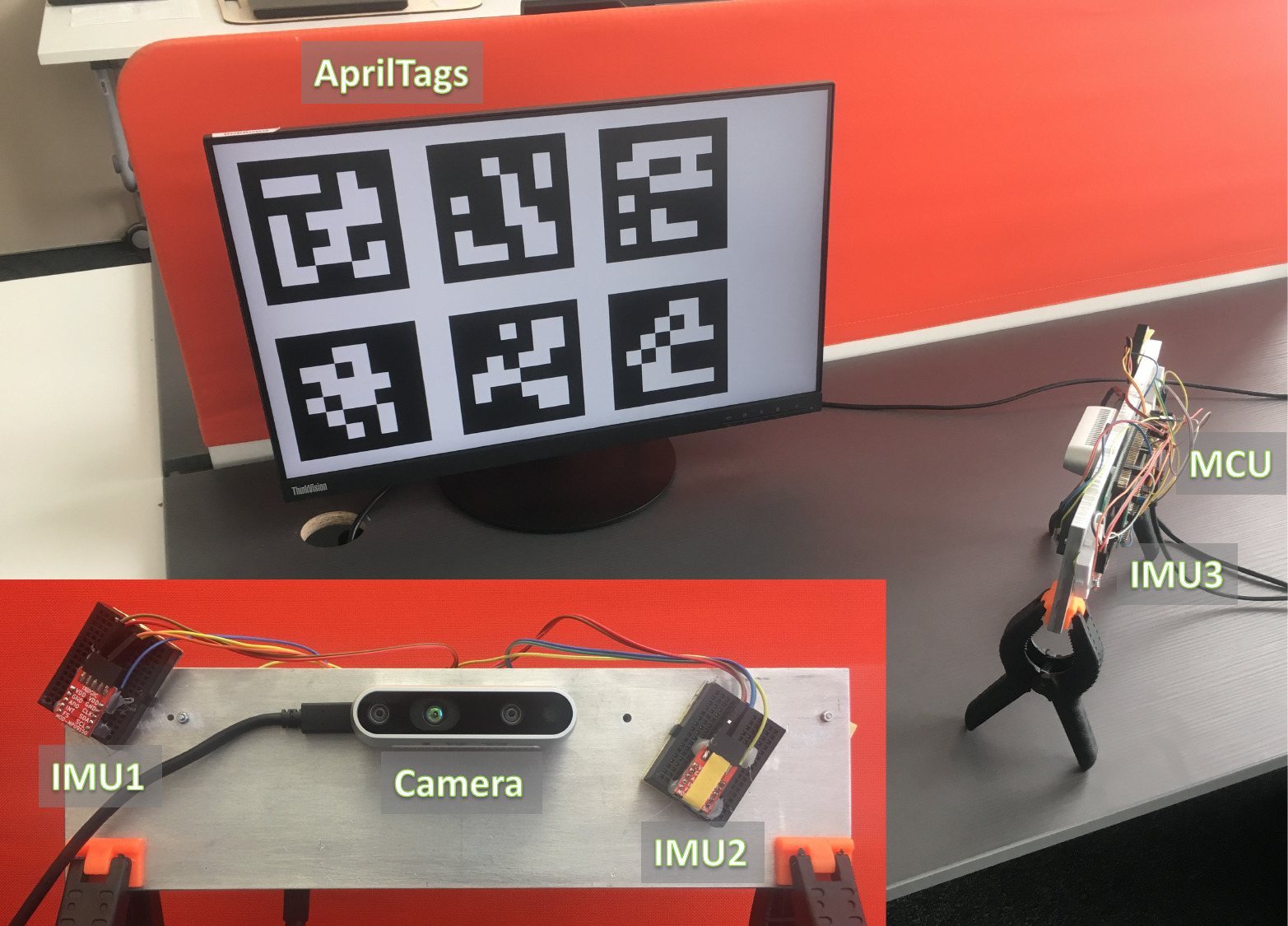}}
\caption{Setup for data collection. Camera and IMUs rigidly connected on aluminium platform. Camera provides ground truth orientation data by detecting pose of motionless AprilTags array displayed on the In-Plane Switching monitor. MCU prepare and transmit IMU data to laptop. AprilTags mean pose coincide with $\{W\}$, IMUi frame is $\{I_i\}$, Camera frame is $\{M\}$.}
\label{fig_sensor}
\end{figure}

We propose to monitor the systematic error and dynamically select among all sensors the best fitting axes to the observed evidence by our proposed algorithm, named {\em Best Axes Composition} (BAC). The underlying hypothesis of BAC is that it is better to discard the information gathered due to the systematic error rather than using every piece of data, which is what a Bayesian approach would do.
This approach of rejecting data could be thought as a statistical hypothesis testing, where we are looking for the most accurate and precise distribution of data, and other distributions, affected differently by the epistemic error, are simply categorized as outlier distributions.
Well-known examples of this approach include RANSAC~\cite{fischler1981random} or the  Minimum Filter~\cite{rfc1059} used in some versions of the Network Time Protocol (NTP).


We will show that empirically this hypothesis holds for the given configurations, and even using a pair of IMUs, we improve the results with respect to other methods for sensor fusion.
We compare our BAC proposed method with two other methods. They are (i) single IMU and (ii) MIMU  Averaged Virtual Estimator (AVE). Single IMU data along with visual information provide comprehensive results in visual-inertial odometry problem~\cite{li2013high, forster2015imu}. MIMU AVE can outperform single IMU methods and shows better accuracy of trajectory estimation~\cite{zhang2019lightweight, bancroft2011data, wang2015}.

The contributions of the paper are:
\begin{itemize}
    \item Best Axes Composition: chooses the most accurate non-coplanar axes among a set of tri-axial IMU sensors for accurate orientation estimation.  
    \item We collected Visual-Inertial data that is utilized for benchmarking BAC. These data include raw video frames and MIMU measurements of rigidly connected sensors as well as ground truth poses of our sensors platform (see Fig.~\ref{fig_sensor}). Data will be released upon acceptance of the present work.
\end{itemize}

\section{Related Work}\label{sec_rl}
The idea of using rigidly-connected Multiple IMU (MIMU) sensors has been investigated for decades,
driven by the idea that multiple measurements from different sensors can improve the accuracy of pose estimation due to the addition of information.
There are many different approaches to fuse sensor data from MIMUs. Nilsson \textit{et al.}~\cite{nilsson2016inertial} state that there are several hundred publications in the field of inertial sensor arrays and methods of MIMU data fusion.
The reasons for using MIMU are extensive and well-grounded: higher accuracy, higher reliability, more accurate uncertainties estimation, higher dynamic measurement range, estimation of angular motion from accelerometer data, direct estimation of angular acceleration, etc. 

The idea of MIMU data fusion is tightly connected with the term virtual IMU that exchanges several real IMUs to the model of single IMU with its own measurements and frame to be used for estimation algorithms. This approach is natural due to the well-defined kinematics of rigidly connected frames expressed by kinematic equations~\cite{bancroft2011data}.

Multiple IMU sensors can be used for failure detection, which is also a widely studied topic. Eckenhoff~\textit{et al.}~\cite{eckenhoff2019sensor} describe a multi-state constraint Kalman filter-based technique to alleviate against IMU sensor failures using switching between the base and auxiliary IMU sensors instead of averaging of measurements. Our work is not directly solving the failure detection problem, although there are many resemblances on the final aim of our approach where we choose the best axes (and discard the rest).

Skog~\textit{et al.}~\cite{skog2014open} introduce an open-source MIMU platform and present error mitigation by averaging out independent stochastic errors. They discuss redundancy and the possibility of alleviating the failure of one of the sensors while the other works fine, different dynamic ranges of sensors can extend the overall dynamic range of fused data when a sensor with lower dynamic range has lower noise density while other has a high dynamic range with the increased value of noise. Skog~\textit{et al.}~\cite{skog2016inertial}~utilize gain in information using geometrical constraints between IMUs on the MIMU platform. Thus, distant accelerometers help in the estimation of angular velocities. Liu~\textit{et al.}~\cite{liu2014design}~and~Edwan~\textit{et al.}~\cite{edwan2011constrained} show gyro-free estimation of angular velocity using four tri-axial and 12 separated mono-axial accelerometers respectively. Zhang~\textit{et al.}~\cite{zhang2019lightweight} utilize up to 9 triaxial IMU sensors to fuse data. The work of Rhudy \textit{et al.}~\cite{rhudy2012fusion} uses three EKF-based fusion methods of GPS while Redundant IMU Data: (a) simple averaging of gyro data before filtering, (b) EKF state prediction step for separate IMUs measurements with update step of averaged state and (c) fully separate filtering with averaging of the state. \cite{xue2012novel} by Xue~\textit{et al.} also solve the problem of noisy gyroscope measurements by a filtering approach. 

While the majority of the publications in the topic, including papers mentioned above, consider a purely probabilistic gyroscope model, Wang~\textit{et al.}~\cite{wang2019optimized} study a more complex model of single-IMU measurements that takes into account non-linear calibration model of measurements such that the model can be considered closer to real sensors. It can be stated that the pure probabilistic model of IMU measurements utilizes only a limited number from a wide range of parameters. For a complete reference in the topic of extending the model of IMU measurements, see \cite{titterton2004strapdown}.

Mass-produced MEMS IMUs became readily available at affordable prices at the expense of marginally deteriorating their accuracy.
Unfortunately, if we require more accuracy for our application, then we might need a high-quality sensor (e.g., fiber optic gyro~\cite{deppe2017mems}), or we could improve the quality of the IMU data by combining several of these cheap sensors. 
Most of the above approaches use a high number of sensors, e.g., Wang and Olson \cite{wang2015} use 72 cheap gyros to provide a highly accurate estimation.
In contrast, we propose to use any number $N$ of gyroscopes, and with as few as 2-3 IMUs, we obtain a good gain with respect to averaging several sensors. The reason is that we treat the systematic error directly while other methods use a large number of sensors hoping to dilute the systematic error.

%% file: 02_background.tex
\section{Background}\label{sec_background}
\subsection{State variables and Frames}\label{sec_frames}

In this paper we use the following frames of reference: the World frame denoted by $\{W\}$, the IMU frame $\{I\}$ or $\{I_i\}$ for each $i$-th IMU in the set of IMUs and the Master frame $\{M\}$ corresponding to an exteroceptive sensor (e.g., visual-camera, depth-camera, lidar, motion capture system, etc.) which is rigidly connected to the frame $\{I\}$. 

We also use the following conventions for expressing vectors. Angular velocities $\lf{I}\vect{\omega} \in \mathbb{R}^3$ 
are gyroscope data, where the frame of reference is denoted with the left-hand superscript $\{I\}$, in this case, the IMU frame.



Rotation matrices $\lfR{\text{Target}}\rf{\text{Origin}} \in SO(3)$ represent a rotation from the frame $\{\text{Origin}\}$ denoted as the right-hand subscript to the frame $\{\text{Target}\}$, as the left-hand superscript. In addition, we express 3D orientations by rotation matrices. 

The rotation $\lfR{W}\rf{I}(t)$ expresses the continuous-time (c-t) orientation of our 3D pose and the rotation matrix $\lfR{W}\rf{I}(t_k)$ is the discrete-time (d-t) sequences of orientations for time $t_k$ and index $k$. The same notation is used for c-t angular velocities $\vect{\omega}(t)$  and d-t $\vect{\omega}(t_k)$.

\subsection{IMU Model}\label{sec_imu}

Measurements of high-end factory-calibrated MEMS gyroscopes can be modeled as~in \cite{nikolic2016characterisation}:
\begin{align}
    \lf{I}\tilde{\vect{\omega}}(t) &= \lf{I}\vect{\omega}(t) + \vect{b}(t) + \vect{\eta}_g(t),
    \label{eq_perf_gyro_cont}
\end{align}
where $\lf{I}\tilde{\vect{\omega}}(t)$ is the measured value by the sensor,  $\lf{I}\vect{\omega}(t)$ the true angular velocity and $\vect{\eta}_g(t)$ is a Wiener process.
The term $\vect{b}(t)$ is the slow varying continuous-time bias modeled as
\begin{align}
    \dot{\vect{b}}(t) = - \frac{1}{\tau_b} \vect{b}(t) + \vect{\epsilon}(t),
    \label{eq_ct_byas}
\end{align}
where $\vect{\epsilon}(t)$ is a Wiener process and
$\tau_b$ is a correlation time of bias~\cite{crassidis2006sigma}.

This model can be further improved by the introduction of a correction lower triangular matrix $C$ to take into account the non-unit scale of measurements and misalignment of axes:
\begin{align}
    \lf{I}\vect{\omega}(t) &= C\,\lf{I}\tilde{\vect{\omega}}(t) - \vect{b}(t) - \vect{\eta}_g(t).
    \label{eq_cor_gyro_cont}
\end{align}

The acceleration influence on the gyroscope model is not included because of its insignificance. It is also assumed that IMU-sensor constitutes by itself a single-point sensor 
within one IMU sensor package. For additional discussion on this, please see \cite{skog2006calibration,rehder2016extending}.

The discrete-time model of (\ref{eq_cor_gyro_cont})  accordingly is
\begin{align}
    \lf{I}\vect{\omega}(t_k) &= C\, \lf{I}\tilde{\vect{\omega}}(t_k) - \vect{b}(t_k) - \vect{n}_g(t_k),
    \label{eq_cor_gyro_dis}
\end{align}
where the continuous-time bias is now  expressed by the discrete-time equation
\begin{align}
    \Delta \vect{b}(t_k) = - \gamma \vect{b}(t_k) + \vect{n}_b(t_k)
    \label{eq_bias_dis}
\end{align}
and $\Delta \vect{b}(t_k)$ is the difference of bias
\begin{equation}
    \Delta \vect{b}(t_k) \triangleq \vect{b}(t_{k+1}) - \vect{b}(t_k).
\end{equation}

The parameter $\gamma$ is a dimensionless coefficient that takes into account bias correlation time as described in the appendix of~\cite{crassidis2006sigma}. The random variable $\vect{n}_b \sim \mathcal{N}(\vect{0}, \Sigma_{\vect{n}_b})$ and $\vect{n}_g \sim \mathcal{N}(\vect{0}, \Sigma_{\vect{n}_g})$.

In this paper, we have chosen a fairly complete gyroscope model, although more complex alternatives are available in the literature \cite{titterton2004strapdown}.
The aim of this work, however, is to accept the limitations of the gyroscope model, where the systematic error is unavoidable, especially for mass-production MEMS IMUs and adapt our algorithm accordingly.
The nature of these errors is out of the scope of this work, however, they may be related to non-linearities, non-Gaussian distributions of noise, additional non-stationary parameters, fabrication tolerances, temperature, discretization errors, external forces, power supply conditions, etc.

\subsection{Kinematic Model}\label{sec_kinem}

The continuous-time kinematic model used in this work, as proposed in \cite{kevin2017modern},  is 
\begin{align}
    [\lf{I}\vect{\omega}]_\times(t) &= \lfR{W}^\top\rf{I}(t) \lfdotR{W}\rf{I}(t),
    \label{eq_corrected_imu}
\end{align}
where our state variable $\lfR{W}\rf{I}(t) \in SO(3)$ is a 3D-orientation represented by a rotation matrix and $[\vect{\omega}]_\times$ is a skew symmetric matrix:
\begin{align}
    [\vect{\omega}]_\times = \begin{bmatrix}  \omega^1\\ \omega^2\\ \omega^3 \end{bmatrix}_\times = \begin{bmatrix}  0 & -\omega^3 & \omega^2\\ \omega^3 & 0 & -\omega^1\\ -\omega^2 & \omega^1 & 0 \end{bmatrix}.
\end{align}
In order to obtain the discrete-time model, we make the following assumption.

{\bf Assumption 1}: {\em Angular velocities $\vect{\omega}$ are piece-wise constant functions over time and their values remain constant in any time interval $[t_k, t_{k+1}]$}.

Alternative models consider continuous-time integration of interpolated measurements by high-order polynomials~\cite{furgale2012continuous}, however, these kinds of approaches are neither exempt of error.


Accordingly, the integration of the angular velocity $\lf{I}\vect{\omega}({t_k})$ at the IMU frame is exactly calculated by
\begin{align}
    \lfR{W}\rf{I}({t_{k+1}}) &= \lfR{W}\rf{I}(t_{k}) \cdot \Exp\left(\lf{I}\vect{\omega}({t_k})\cdot \Delta t_k\right) \label{eq_irots_dis},
\end{align}
where $\Delta t_k = t_{k+1}-t_k$ and the function $\Exp(\vect{\theta})$ is a mapping from the tangent space of rotations $\mathbb{R}^3$ (spanning the Lie algebra $\mathfrak{so}(3)$) to the group of matrix rotations $SO(3)$. The inverse operation exists, the logarithm $\Ln:~SO(3)~\to~\mathbb{R}^3$.
The topic of Lie algebra is well documented and it is a convenient way to express rotations or rigid body transformations \cite{kevin2017modern, barfoot2017}.



The orientation of the Master frame $\{M\}$, given any of the $i$-th IMUs can be computed by the similar equation:
\begin{align}
    \lfR{W}^i\rf{M}({t_{k+1}}) &= \lfR{W}^i\rf{M}(t_{k}) \cdot \Exp\left(\lf{M}\vect{\omega}({t_k})\cdot \Delta t_k\right),
    \label{eq_mrots_dis}
\end{align}
where the angular velocity expressed via angular velocity of some of gyroscopes by
\begin{align}\label{eq_omega_rot}
    \lf{M}\vect{\omega} = \lfR{M}\rf{I_i}\lf{I_i}\vect{\omega}.
\end{align}

\subsection{Averaged Virtual Estimator}\label{sub_baseline}

Given multiple gyroscope sensor readings, one can estimate overall orientation by fusing all available data into a virtual angular velocity:
\begin{align}
    \lf{V}\hat{\vect{\omega}} = \frac{1}{N}\sum_{i=1}^N \lfR{V}\rf{I_i} \left(\lf{I_i}\tilde{\vect{\omega}} - \hat{\vect{b}}_{i}\right),
    \label{eq_baseline}
\end{align}
where $N$ is the number of IMUs and $\lfR{V}\rf{I_i}$ is the rotation to transform IMU measurements to the virtual frame $\{V\}$ under assumption of independent noise processes with the same parameters in~(\ref{eq_perf_gyro_cont})~and~(\ref{eq_ct_byas}), matrix $C$ is not shown.

We have chosen to compare against one variant of the {\em virtual} IMU approach \cite{zhang2019lightweight}.
The authors utilize a probabilistic version of (\ref{eq_perf_gyro_cont}).
The authors report  that virtual IMU measurements constructed by the weighted sum of measurements from multiple IMUs gives better accuracy compared to single-IMU measurements. They also consider identical IMUs that simplify the measurement model to the expectation of measurements. We refer to the approach in (\ref{eq_baseline}) as the  Averaged Virtual Estimator (AVE) and will use this model as the baseline for the comparison with our proposed model.

%% file: 03_method.tex
\section{Best Axes Composition for Multiple IMU Sensor Fusion in Open Loop}\label{sec_approach}

If the proposed IMU model (\ref{eq_cor_gyro_dis}) truthfully supports real data, that is, the statistical error has been accurately estimated, then the more data we use, the more accurate will be the estimator, ultimately leading to an averaged estimator as in (\ref{eq_baseline}) or similar approaches.

On the contrary, one can argue that the previous statement does not always hold, especially for mass-production IMUs, and we must make a different assumption.

{\bf Assumption 2: }{\em There exists a non-negligible source of the error, which can not be quantified in the current model (systematic uncertainty), which affects each axis of each IMU differently at each instant of time.}

One could calculate the statistics of this error given ground truth information. However, this will lead to more complex models, and still, they will not be exempt from systematic error.

We propose a different approach to fuse sensor data, which is based on the following idea. 

{\bf Hypothesis: }{\em The systematic error at each axis can not be predicted; however, it can be evaluated by the error in the estimated state. This error is non-stationary but does not change significantly for short time intervals.}

Based on this, we propose to choose dynamically those IMU axes which have presented minimum error during evaluation in a method called {\em Best Axes Composition} (BAC).
In Sec.~\ref{sec_exp}, we show empirically the validity of the Hypothesis for the dataset configuration used.


We propose three stages or modes of working: (A) stationary and (B) time-dependent IMU parameters estimation when reference is observed (closed loop); (C) open loop state estimation with no observations from reference.

\subsection{Stationary IMU Parameters Estimation}\label{sec_static_optim}

In order to evaluate all the available IMUs we require an exteroceptive sensor (at the Master frame $\{M\}$) to provide ``ground truth'' measurements of absolute orientation compared with the estimated orientation from each IMU.

In the first preliminary stage we perform batch optimization for estimating the extrinsic and intrinsic parameters of IMUs, namely scale-misalignment matrix $C_i$ introduced in~(\ref{eq_cor_gyro_cont}) and rotations of IMUs w.r.t.~Master $\lfR{M}\rf{I_i}$. Time-depended biases and orientations are also estimated. This stage has an important role and will be discussed in~\ref{sec_exp}.
States and parameters to be estimated are
\begin{align}
    \Theta_i = \{ \Ln(\lfR{W}^{i}\rf{M}(t_k)), \vect{b}^i(t_k), C_i, \Ln(\lfR{M}\rf{I_i})\}.
\end{align}
The number of the variables is $(6K + 9)N$ for $N$ IMU sensors and $K$ timestamps.

Assuming eqs.~(\ref{eq_cor_gyro_dis}),~(\ref{eq_bias_dis}),~(\ref{eq_mrots_dis}),~(\ref{eq_omega_rot}), the cost function given the values is the following for each IMU
\begin{align*}
    \mathcal{L}_i(\Theta_i) =& \sum_{k=0}^K \left\| \Ln (\lfR{M}^{\scriptscriptstyle GT}\rf{W}(t_k) \lfR{W}^i\rf{M}(t_k)) \right\|_{\Sigma_{\theta}}^2\\
    +& \sum_{k=0}^{K-1} \left\| \Delta \vect{b}^i(t_k) + \gamma \vect{b}^i(t_k)\right\|_{\Sigma_{\vect{n}_b}}^2, \numberthis
    \label{eq_loss}
\end{align*}
where the first term is the error in orientation and the second term is condition to bias, $\lfR{M}^{\scriptscriptstyle GT}\rf{W}(t_k)$ is ground truth data provided by Master sensor, $\lfR{W}^i\rf{M}(t_k)$ is orientation estimate, 
$\Sigma_\theta$ is a covariance matrix described in~\cite{forster2015supplementary}~and~\cite{sola2018micro}.

The final cost of all IMU measurements to be minimized is
\begin{align}
    \mathcal{L}(\Theta) =& \sum_{i=0}^N \mathcal{L}_i(\Theta_i) \to \min_{\Theta}.
    \label{eq_final_loss}
\end{align}

\subsection{Time-dependent IMU Parameters Estimation}

Second stage is the initial part of the main algorithm. Again, we treat observations from the camera as ground truth orientations. It includes further continuous optimization of the cost in (\ref{eq_final_loss}) but with 
only states to be estimated: $\Theta_i~=~ \{\Ln(\lfR{W}^{i}\rf{M}(t_k)), \vect{b}^i(t_k)\}$, $6KN$ values in total. Here we use formerly estimated scale-misalignment, rotation and constant bias parameters for every IMU. 


Estimation of biases in this stage is followed by the best axes choice for the third stage algorithm where orientation estimation is performed in open loop without Master's aid.

\subsection{Best Axes Composition}\label{sec_bac}

Our approach, the Best Axes Composition (BAC), performs choice of three non-coplanar axes to be utilized from the whole set of $3N$ axes for open-loop orientation estimation instead of averaging, as described in~\ref{sub_baseline}. Other axes are ignored for data fusion since they are worse fitted and thus, the systematic error should be bigger. The criterion for the BAC to choose specific axis among the set of the homonymous (but not necessarily collinear) axes of different IMUs is: \textit{the lowest estimated mean squared error within the last $p$ Master aided measurements}. 

Below is the description of the BAC approach.

There are $p$ last reference $\lfR{W}^{\scriptscriptstyle GT}\rf{M}(t_k)$ and estimated $\lfR{W}^M\rf{M}(t_k)$ orientations of the Master of the second stage. 
The orientation error for IMU $i$ and timestamp $k$ expressed in the respective IMU frame~$i$ is 
\begin{align*}
    \vect{e}_{i}(t_k) =& \Ln\left(\left( \lfR{W}^{\scriptscriptstyle GT}\rf{M}(t_k) \lfR{M}\rf{I_i}\right)^\top \lfR{W}^i\rf{M}(t_k) \lfR{M}\rf{I_i}\right)\\
    =& \Ln\left( \lfR{I_i}\rf{M} \lfR{M}^{\scriptscriptstyle GT}\rf{W}(t_k) \lfR{W}^i\rf{M}(t_k) \lfR{M}\rf{I_i} \right). \numberthis
\end{align*}

Here, the mapping of the error in the Master to the IMU frames has been performed.
The mapping allows the algorithm to capture the contribution of measurements of each single axis of IMU to three-dimensional error vector 
\begin{align}
    \vect{e}_{i} = [e^x_i, e^y_i, e^z_i]^\top.
\end{align}

Then we choose the best axis from the homonymous axes of the IMUs. This is done by MSE criterion of the sets of errors $e^\alpha_i$ in time instances $t_{K-p}$,~$t_{K-p+1}$,~\dots,~$t_{K}$ for every axis $\alpha~\in \{x,y,z\}$ ($\alpha~\in \{1,2,3\}$ is used interchangeably) with the assumption that chosen axes are not coplanar:
\begin{align}
    i_{best}^\alpha = \arg \min_i \sum_{k=K-p}^K | e^\alpha_i (t_k)|^2
    \label{eq_vot_tule}
\end{align}

We do it three times for every axis to compose axes of the virtual IMU. Thus, on this step the set of three non-coplanar axes \begin{align}
    \{i_{best}^x, i_{best}^y, i_{best}^z\}
    \label{eq_best_axes}
\end{align}
from the set of IMU sensors for open-loop estimation have been chosen by criterion (\ref{eq_vot_tule}). 

The next step is the final composition of the axes and their measurements to assemble accurate distributed single virtual IMU measurements. We match axes of the virtual IMU with the Master frame for simplicity, but there are no limitations on the position and orientation of the virtual IMU. 
For this, consider now a problem of expression of angular velocity in the Master frame via non-coplanar elements of angular velocities 
$\lf{I_i}\omega^x,~\lf{I_j}\omega^y$ and $\lf{I_k}\omega^z$ 
of arbitrary chosen IMUs~$i$,~$j$~and~$k$:
\begin{align}\label{eq_map}
    \lf{M}\vect{\omega} = A \cdot \begin{bmatrix} \lf{I_i}\omega^x\\ \lf{I_j}\omega^y\\ \lf{I_k}\omega^z \end{bmatrix},
\end{align}
where $A \in \mathbb{R}^{3\times3}$ is a linear mapping that is needed to be found. The matrix $A$ is non-orthonormal and does not belong to $SO(3)$ because the homonymous axes of different IMUs do not have to be coplanar in general. The inverse of $A$ can be obtained by exploring element by element of the  vector
\begin{align}
    \lf{I_i}\vect{\omega} = \lfR{I_i}\rf{M}\lf{M}\vect{\omega}
\end{align}
which is the inverse relation in (\ref{eq_omega_rot}). This gives each component of the angular velocity as a dot product of
\begin{align}\label{eq_comp}
    \lf{I_i}\omega^\alpha = \lf{I_i} \vect{r}^\alpha \rf{M} \lf{M}\vect{\omega},
\end{align}
for every axis $\alpha$ of $\lf{I_i}\vect{\omega}$, where the row vector $\vect{r}^\alpha \rf{M}$ is the $\alpha$-th row of $\lfR{I_i}\rf{M}$. 

By stacking all the velocity components we obtain
\begin{align}
    \begin{bmatrix} \lf{I_i}\omega^x\\ \lf{I_j}\omega^y\\ \lf{I_k}\omega^z \end{bmatrix} = \begin{bmatrix} \lf{I_i} \vect{r}^x \rf{M}\\ \lf{I_j} \vect{r}^y \rf{M}\\ \lf{I_k} \vect{r}^z \rf{M} \end{bmatrix} \lf{M}\vect{\omega}.
\end{align}
and finally,
\begin{align}\label{eq_distr}
    \lf{M}\vect{\omega} = \begin{bmatrix} 
        \lf{I_i} \vect{r}^x \rf{M}\\ 
        \lf{I_j} \vect{r}^y \rf{M}\\ 
        \lf{I_k} \vect{r}^z \rf{M}
    \end{bmatrix}^{-1} \begin{bmatrix} \lf{I_i}\omega^x\\ \lf{I_j}\omega^y\\ \lf{I_k}\omega^z \end{bmatrix}
\end{align}
is the solution of~(\ref{eq_map}).

Changing superscripts $I_i$, $I_j$, $I_k$ in~(\ref{eq_distr}) on $I_{i_{best}^x}, I_{i_{best}^y}, I_{i_{best}^z}$
accordingly we get BAC. As a result for the third stage, we have prepared accurate distributed IMU (\ref{eq_distr}) for open-loop orientation estimation. 

\subsection{Open-loop}
The third stage of the approach is open-loop orientation estimation by only IMUs measurements without Master aid. This is an integration of distributed IMU measurements (\ref{eq_distr}) with constant bias equal to the last estimated value in the second stage of the approach $\vect{b} = \vect{b}(t_K)$. 

As was mentioned in the Section~\ref{sec_intro}, highly dynamic motion of aerial robots or augmented and virtual reality systems require as low as possible processing time of data along with frequent updates of current state (high data rate). Using only visual data requires several tens or hundreds of milliseconds delay between events and their estimation. In these circumstances IMU sensors provide a much faster alternative and BAC outperforms state-of-the-art IMU solutions in accuracy over short time horizons. 

%% file: 04_experiments.tex
\section{Implementation}

In order to carry out experiments and evaluate our approach, we developed a handheld visual-inertial sensor consisting of three IMUs MPU9150, Intel RealSense D435 depth camera (only one gray-scale global-shutter visual camera was used) calibrated beforehand and a developers board based on the MCU STM32F051.
All the sensors are rigidly placed on the metal platform as shown in Fig.~\ref{fig_sensor}.

Ground truth data are provided by tracking a custom designed array of six AprilTags~\cite{olson2011apriltag}. This array is needed in order to check for the mutual error between the poses of every detected tag for robustness on the ground truth data generated.
In addition, we use a screen to show the tags instead of a printed paper sheet to ensure that tags present no distortions and they are still valid fiducial markers since dimensions are known by the pixel size of the monitor. The World frame $\{W\}$ coincides with the frame of this array of markers, averaging of the orientation $\lfR{W}\rf{M}$ performed by averaging in $SO(3)$~\cite{moakher2002means}.

The MCU performs data collection of IMU data, all the IMU sensors are synchronized on the hardware level and have a highly accurate quartz clock generator. 
Data rate of synchronized IMUs is 342 Hz. Frame rate of the camera is 30 fps with constant exposure time. Time synchronization Camera-IMUs is performed by correlation of absolute value of angular velocities, and has accuracy better than 3 ms.

IMUs are manually placed on the board roughly maintaining the same orientations. It is enough to avoid any singularities explained in Sec.~\ref{sec_bac}.

Batch optimization is performed in PyTorch Python package by optimizing on the manifold of rotation matrices according to the technique described in~\cite{forster2015imu}. The chosen optimizer is Adam with variable learning rate 0.001-0.1 and 1400 epochs. 

Ground truth orientation was interpolated by a cubic spline to be coincided with every IMU measurements. We have used the same prior covariance for each IMU axis, although in the literate some authors make a distinction regarding the z-axis \cite{nikolic2016characterisation}.



\section{Experiments}\label{sec_exp}
For evaluation we gathered 44 non-overlapping tracks, each lasting 15 sec in duration: 10 sec and 5 sec for orientation estimation with Master aid and in open loop respectively.

The first stage of the experiments was carried out separately with a specific trajectory. It was important because common tracks have pure diversity to get accurate estimations for constant parameters of the sensor. 
This stage can also be done by other calibration tools, for instance, by~\cite{rehder2016extending}.

We consider a comparison with the competitors in short horizons of time that last 5 s to highlight (i) the dynamic nature of BAC and (ii) the need of proper treatment of this method. We state that the method greatly works right after loss of complement sensor (camera, lidar, etc.) data and degrades in a higher horizon of time to MIMU or single IMU case and needs repetitive updates by aiding sensor data to outperform the competitors continuously.

\begin{figure}[t]
\includegraphics[width=\columnwidth]{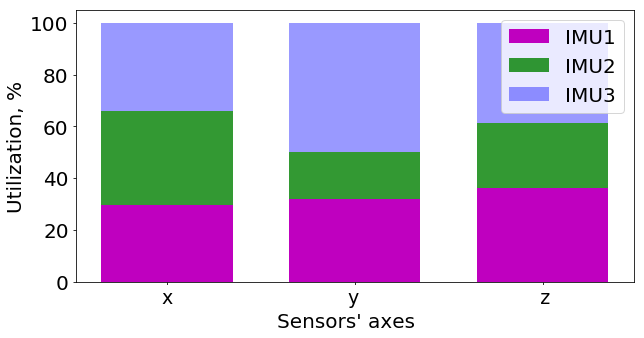}
\caption{Frequency of utilization of every axis of every IMU-sensor. This figure shows that the selection of the best axes is time varying.}
\label{fig_freq}
\end{figure}

Fig.~\ref{fig_freq} shows that the proposed method dynamically utilizes every axis of every IMU during the experiments, confirming empirically that the nature of the error is non-stationary and different for each IMU.

\begin{figure}[t]
\includegraphics[width=0.95\columnwidth]{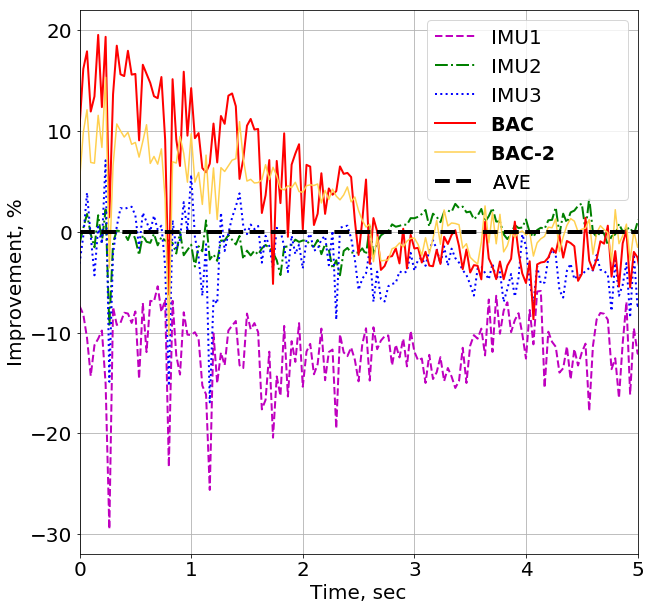}
\caption{Accuracy comparison for the stage III. Orientation accuracy improvement of our proposed method wrt. AVE averaging method in percent for 44 tracks averaged value of errors. The higher curve, the better accuracy. Single IMUs estimation accuracy also depicted.}
\label{fig_impr}
\end{figure}

We have compared the averaged ratio (in percent) between the chosen baseline AVE and other estimation methods: single IMU and our proposed method in Fig.~\ref{fig_impr}. 
The higher the curve the better accuracy of a method compared with the AVE, which is a flat constant line. It can be seen, our proposed BAC method outperforms significantly the baseline up to 2.5 sec in open loop and shows more than 10\% better performance within one second time horizon. 

The results also shows the same performance of IMU2 and IMU3 of the sensor while pure performance of IMU1 in average. 

We also applied our approach for two IMU sensors (BAC-2), namely IMU2 and IMU3. While accuracy for this setup is less then BAC using three IMUs, it is clear that it is also better than three IMU AVE within the same horizon. This sugests the idea that even less accurate sensor (IMU1 in our case) contribute well to overall improvement.

\begin{figure}[t]
\includegraphics[width=\columnwidth]{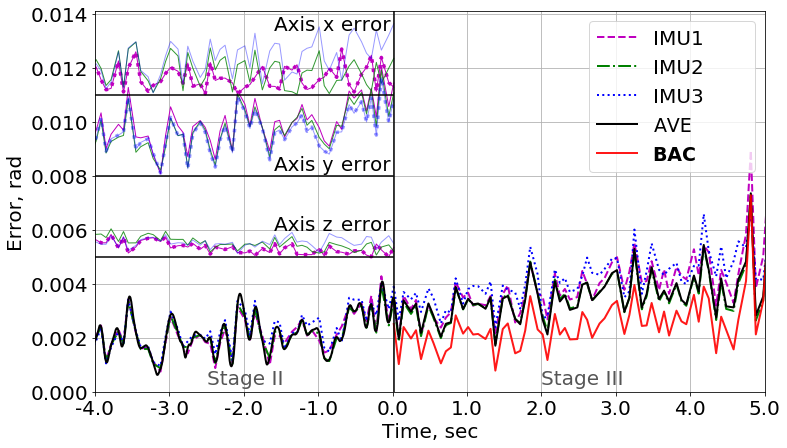}
\caption{Typical estimation error with Master aid (Stage II) and in open-loop (Stage III). Top left part is error in orientation measured for every axis of every IMU sensor in the stage II, dotted curves are chosen by proposed algorithm for Stage III.}
\label{fig_2-3}
\end{figure}

Fig.~\ref{fig_2-3} expresses the typical error value of the second and third stages for different IMUs and methods. On the left top corner of the picture error values by every axis of IMUs are depicted. There curves marked by bold are chosen as more accurate by our BAC criteria. We can see that these chosen axes for estimation of orientation in the third stage gives better accuracy than every single IMU or AVE data fusion method and stays lower than 0.004 rad in 5 seconds time horizon. 

%% file: 05_conclusion.tex
\section{Conclusions}\label{sec_concl}

We have proposed the Best Axes Composition method of IMU sensors data fusion that takes into account systematic error  by choosing three non-coplanar axes from different IMU sensors. To do that, we have proposed and verified the Hypothesis and as a result, our approach outperforms the Averaged Virtual Estimator method up to several seconds of open loop and further accuracy of both methods remains similar. This result is reasonable to think since we can only estimate the systematic error for short time periods.
We have shown that our method requires only two IMUs to significantly improve orientation estimation.

Further work can be directed to a combination of our proposed axes-choice and more widely used averaging approaches in order to treat the best sides of both probabilistic and systematic error assumptions. However, the first step of the further development is to construct an entire pipeline on top of the method for online functioning. This is possible due to the absence of any time-consuming operation.
